\documentclass[10pt,twocolumn,letterpaper]{article}

\usepackage[pagenumbers]{cvpr} 

\usepackage{graphicx}
\usepackage{amsmath}
\usepackage{amssymb}
\usepackage{booktabs}

\usepackage{multirow}
\usepackage{makecell}

\usepackage{algorithm}
\usepackage{algorithmic}

\usepackage{amsfonts}
\usepackage{setspace}
\usepackage{multirow}
\usepackage{diagbox}
\DeclareMathOperator*{\argmax}{argmax}
\makeatletter
\newcommand*\bigcdot{\mathpalette\bigcdot@{.5}}
\newcommand*\bigcdot@[2]{\mathbin{\vcenter{\hbox{\scalebox{#2}{$\m@th#1\bullet$}}}}}

\usepackage[pagebackref,breaklinks,colorlinks]{hyperref}

\usepackage[capitalize]{cleveref}
\crefname{section}{Sec.}{Secs.}
\Crefname{section}{Section}{Sections}
\Crefname{table}{Table}{Tables}
\crefname{table}{Tab.}{Tabs.}

\def\cvprPaperID{9950} 
\def\confName{CVPR}
\def\confYear{2022}

\begin{document}

\title{AutoMC: Automated Model Compression based on Domain  Knowledge and Progressive search strategy}

\author{Chunnan Wang, Hongzhi Wang, Xiangyu Shi\\
Harbin Institute of Technology \\
{\tt\small \{WangChunnan,wangzh,xyu.shi\}@hit.edu.cn}
}
\maketitle

\begin{abstract}
Model compression methods can reduce model complexity on the premise of maintaining acceptable performance, and thus promote the application of deep neural networks under resource constrained environments. Despite their great success, the selection of suitable compression methods and design of details of the compression scheme are difficult, requiring lots of domain knowledge as support, which is not friendly to non-expert users. To make more users easily access to the model compression scheme that best meet their needs, in this paper, we propose AutoMC, an effective automatic tool for model compression. AutoMC builds the domain knowledge on model compression to deeply understand the characteristics and advantages of each compression method under different settings. In addition, it presents a progressive search strategy to efficiently explore pareto optimal compression scheme according to the learned prior knowledge combined with the historical evaluation information. Extensive experimental results show that AutoMC can provide satisfying compression schemes within short time, demonstrating the effectiveness of AutoMC.
\end{abstract} 


\section{Introduction}\label{section:1}

Neural networks are very powerful and can handle many real-world tasks, but their parameter amounts are generally very large bring expensive computation and storage cost. In order to apply them to mobile devices building more intelligent mobile devices, many model compression methods have been proposed, including model pruning~\cite{C2,C3,C4,C5,C6}, knowledge distillation~\cite{C1}, low rank approximation~\cite{C5,C7} and so on.

These compression methods can effectively reduce model parameters while maintaining model accuracy as much as possible, but are difficult to use. Each method has many hyperparameters that can affect its compression effect, and different methods may suit for different compression tasks. Even the domain experts need lots of time to test and analyze for designing a reasonable compression scheme for a given compression task. This brings great challenges to the practical application of compression techniques. 

In order to enable ordinary users to easily and effectively use the existing model compression techniques, in this paper, we propose AutoMC, an \underline{Auto}matic \underline{M}achine \underline{L}earning (AutoML) algorithm to help users automatically design model compression schemes. Note that in AutoMC, we do not limit a compression scheme to only use a compression method under a specific setting. Instead, we allow different compression methods and methods under different hyperparameters settings to work together (execute sequentially) to obtain diversified compression schemes. We try to integrate advantages of different methods/settings through this sequential combination so as to obtain more powerful compression effect, and our final experimental results prove this idea to be effective and feasible.

However, the search space of AutoMC is huge. The number of compression strategies\footnote{In this paper, a compression strategy refers to a compression method with a specific hyperparameter setting.} contained in the compression scheme may be of any size, which brings great challenges to the subsequent search tasks. In order to improve the search efficiency, we present the following two innovations to improve the performance of AutoMC from the perspectives of knowledge introduction and search space reduction, respectively. 

Specifically, for the first innovation, we built domain knowledge on model compression, which discloses the technical and settings details of compression strategies, and their performance under some common compression tasks. This domain knowledge can assist AutoMC to deeply understand the potential characteristics and advantages of each component in the search space. It can guide AutoMC select more appropriate compression strategies to build effective compression schemes, and thus reduce useless evaluation and improve the search efficiency.

As for the second innovation, we adopted the idea of progressive search space expansion to improve the search efficiency of AutoMC. Specifically, in each round of optimization, we only take the next operations, i.e., unexplored next-step compression strategies, of the evaluated compression scheme as the search space. Then, we select the pareto optimal operations for scheme evaluation, and finally take the next operations of the new scheme as the newly expanded search area to participate in the next round of optimization. In this way, AutoMC can selectively and gradually explore more valuable search space, reduce the search difficulty, and improve the search efficiency. In addition, AutoMC can analyze and compare the impact of subsequent operations on the performance of each compression scheme in a fine-grained manner, and finalize a more valuable next-step exploration route for implementation, thereby effectively reducing the evaluation of useless schemes.

The final experimental results show that AutoMC can quickly search for powerful model compression schemes. Compared with the existing AutoML algorithms which are non-progressive and ignore domain knowledge, AutoMC is more suitable for dealing with the automatic model compression problem where search space is huge and components are complete and executable algorithms.

Our contributions are summarized as follows:
\begin{itemize}
\item [1. ] \textbf{Automation.} AutoMC can automatically design the effective model compression scheme according to the user demands. As far as we know, this is the first automatic model compression tool. 
\vspace{-0.1cm}
\item [2. ] \textbf{Innovation.} In order to improve the search efficiency of AutoMC algorithm, an effective analysis method based on domain knowledge and a progressive search strategy are designed. As far as we know, AutoMC is the first AutoML algorithm that introduce external knowledge.
\vspace{-0.1cm}
\item [3. ] \textbf{Effectiveness.} Extensive experimental results show that with the help of domain knowledge and progressive search strategy, AutoMC can efficiently search the optimal model compression scheme for users, outperforming compression methods designed by humans.
\end{itemize}

\section{Related Work}\label{section:2}
\subsection{Model Compression Methods}

Model compression is the key point of applying neural networks to mobile or embedding devices, and has been widely studied all over the world. Researchers have proposed many effective compression methods, and they can be roughly divided into the following four categories. (1) pruning methods, which aim to remove redundant parts e.g., filters, channels, kernels or layers, from the neural network~\cite{p5,p6,p7,p9}; (2) knowledge distillation methods that train the compact and computationally efficient neural model with the supervision from well-trained larger models; (3) low-rank approximation methods that split the convolutional matrices into small ones using decomposition techniques~\cite{p4}; (4) quantization methods that reduce the precision of parameter values of the neural network~\cite{p8,p20}. 

These compression methods have their own advantages, and have achieved great success in many compression tasks, but are difficult to apply as is discussed in the introduction part. In this paper, we aim to flexibly use the experience provided by them to support the automatic design of model compression schemes.

\subsection{Automated Machine Learning Algorithms}
 
The goal of \underline{Auto}mated \underline{M}achine \underline{L}earning (AutoML) is to realize the progressive automation of ML, including automatic design of neural network architecture, ML workflow~\cite{p18,p19} and automatic setting of hyperparameters of ML model~\cite{p16,p17}. The idea of the existing AutoML algorithms is to define an effective search space which contains a variety of solutions, then design an efficient search strategy to  quickly find the best ML solution from the search space, and finally take the best solution as the final output.

Search strategy has a great impact on the performance of the AutoML algorithm. The existing AutoML search strategies can be divided into 3 categories: \underline{R}einforcement \underline{L}earning (RL) methods~\cite{p10}, \underline{E}volutionary \underline{A}lgorithm (EA) based methods~\cite{p11,p12} and gradient-based methods~\cite{p13,p14}. The RL-based methods use a recurrent network as controller to determine a sequence of operators, thus construct the ML solution sequentially. EA-based methods initialize a population of ML solutions first and then evolve them with their validation accuracies as fitnesses. As for the gradient-based methods, they are designed for neural architecture search problems. They relax the search space to be continuous, so that the architecture can be optimized with respect to its validation performance by gradient descent~\cite{p15}. They fail to deal with the search space composed of executable compression strategies. Therefore, we only compare AutoMC's search strategy with the previous two methods.

\begin{table*}[t]
\caption{Six open source compression methods that are used in our search space. $\ast n$ denotes multiply $n$ by the number of pre-training epochs of the original model $M$, and $HP_2=\times \gamma$ means reduce $P(M)\times \gamma$ parameters from $M$.}
\vspace{-0.3cm}
\small
\newcommand{\tabincell}[2]{\begin{tabular}{@{}#1@{}}#2\end{tabular}}
\centering
\resizebox{0.83\textwidth}{!}{
\smallskip\begin{tabular}{m{0.8cm}|m{1.8cm}|m{5.0cm}|m{10.5cm}}
\hline
\textbf{Label} & \textbf{Compression Method} & \textbf{Techniques} & \textbf{Hyperparameters} \\
\hline
C1 & LMA ~\cite{C1} & $TE_1$: Knowledge distillation based on LMA function & \tabincell{l}{$\bigcdot$ $HP_1: $ fine tune epochs $\in$ $\{*0.1, *0.2, *0.3, *0.4, *0.5\}$ \\ $\bigcdot$ $HP_2: $ decrease ratio of parameter $\in$ $\{\times0.04, \times0.12, \times0.2, \times0.36, \times0.4\}$ \\  $\bigcdot$ $HP_3: $ LMA's segment number $\in$ $\{6, 8, 10\}$ \\ $\bigcdot$ $HP_4: $ temperature factor $\in$ $\{1, 3, 6, 10\}$ \\ $\bigcdot$ $HP_5: $ alpha factor $\in$ $\{0.05, 0.3, 0.5, 0.99\}$ } \\
\hline
C2 & LeGR ~\cite{C2} & \tabincell{l}{$TE_2$: Filter pruning based on EA \\ $TE_3$: Fine tune} & \tabincell{l}{$\bigcdot$ $HP1, HP2: $ same as that in C1 \\  $\bigcdot$ $HP_6: $ channel's maximum pruning ratio $\in$ $\{0.7, 0.9\}$ \\ $\bigcdot$ $HP_7: $ evolution epochs $\in$ $\{*0.4, *0.5, *0.6, *0.7\}$ \\ $\bigcdot$ $HP_8: $ filter’s evaluation criteria $\in \{l1\_weight, l2\_weight, l2\_bn, l2\_bn\_param\}$ } \\
\hline
C3 & NS ~\cite{C3} & \tabincell{l}{$TE_4$: Channel pruning based on Scaling \\ Factors in BN Layers \\ $TE_3$: Fine tune} & \tabincell{l}{$\bigcdot$ $HP_1, HP_2: $ same as that in C1 \\  $\bigcdot$ $HP_6: $ same as that in C2} \\
\hline
C4 & SFP ~\cite{C4} & $TE_5$:  Filter pruning based on back-propagation & \tabincell{l}{$\bigcdot$ $HP_2: $ same as that in C1 \\  $\bigcdot$ $HP_9: $ back-propagation epochs $\in$ $\{*0.1, *0.2, *0.3, *0.4, *0.5\}$ \\ $\bigcdot$ $HP_{10}: $ update frequency $\in$ $\{1, 3, 5\}$} \\
\hline
C5 & HOS ~\cite{C5} & \tabincell{l}{$TE_6$:  Filter pruning based on HOS\\ \cite{HOS} \\ $TE_7$: Low-rank kernel approximation\\ based on HOOI~\cite{HOOI} \\ $TE_3$: Fine tune} & \tabincell{l}{$\bigcdot$ $HP_1, HP_2: $ same as that in C1 \\  $\bigcdot$ $HP_{11}: $ global evaluation criteria $\in$ $\{P1, P2, P3\}$ \\ $\bigcdot$ $HP_{12}: $ global evaluation criteria $\in$ $\{l1norm, k34, skew\_kur\}$ \\ $\bigcdot$ $HP_{13}: $ optimization epochs $\in$ $\{*0.3, *0.4, *0.5\}$ \\ $\bigcdot$ $HP_{14}: $ MSE loss's factor $\in$ $\{1, 3, 5\}$} \\
\hline
C6 & LFB ~\cite{C7} & $TE_9$:  low-rank filter approximation based on filter basis & \tabincell{l}{$\bigcdot$ $HP_1, HP_2: $ same as that in C1 \\  $\bigcdot$ $HP_{15}: $ auxiliary MSE loss's factor $\in$ $\{0.5, 1, 1.5, 3, 5\}$ \\ $\bigcdot$ $HP_{16}: $ auxiliary loss $\in$ $\{NLL, CE, MSE\}$} \\
\hline
\end{tabular}
}
\label{table1}
\vspace{-0.4cm}
\end{table*}

\section{Our Approach}\label{section:3}
We firstly give the related concepts on model compression and problem definition of automatic model compression (Section~\ref{section:3.1}). Then, we make full use of the existing experience to construct an efficient search space for the compression area (Section~\ref{section:3.2}). Finally, we designed a search strategy, which improves the search efficiency from the perspectives of knowledge introduction and search space reduction, to help users quickly search for the optimal compression scheme (Section~\ref{section:3.3}).

\subsection{Related Concepts and Problem Definition}\label{section:3.1}
\textbf{Related Concepts.} Given a neural model $M$, we use $P(M)$, $F(M)$ and $A(M)$ to denote its parameter amount, FLOPS and its accuracy score on the given dataset, respectively. Given a model compression scheme $S=\{s_{1}\to s_{2}\to \ldots \to s_{k}\}$, where $s_{i}$ is a compression strategy ($k$ compression strategies are required to be executed in sequence), we use $S[M]$ to denote the compressed model obtained after applying $S$ to $M$. In addition, we use $*R(S,M)=\frac{*(M)-*(S[M])}{*(M)}\in [0,1]$, where $*$ can be $P$ or $F$, to represent model $M$’s reduction rate on parameter amount or FLOPS after executing $S$. We use $AR(S,M)=\frac{A(S[M])-A(M)}{A(M)} >-1$ to represent accuracy increase rate achieved by $S$ on $M$.

\textbf{Definition 1 (Automatic Model Compression).} Given a neural model $M$, a target reduction rate of parameters $\gamma$ and a search space $\mathbb{S}$ on compression schemes, the Automatic Model Compression problem aims to quickly find $S^{\ast}\in \mathbb{S}$:
\begin{equation}
\small
\begin{split}
S^{\ast}&=\mathop{\argmax}_{S\in \mathbb{S},PR(S,M)
geq \gamma} f(S,M) \\
f(S,&M):=[AR(S,M),PR(S,M)]
\end{split}
\vspace{-0.2cm}
\end{equation}
A Pareto optimal compression scheme that performs well on two optimization objectives: $PR$ and $AR$, and meets the target reduction rate of parameters.

\begin{figure}[t]
\centering
\includegraphics[width=0.9\columnwidth]{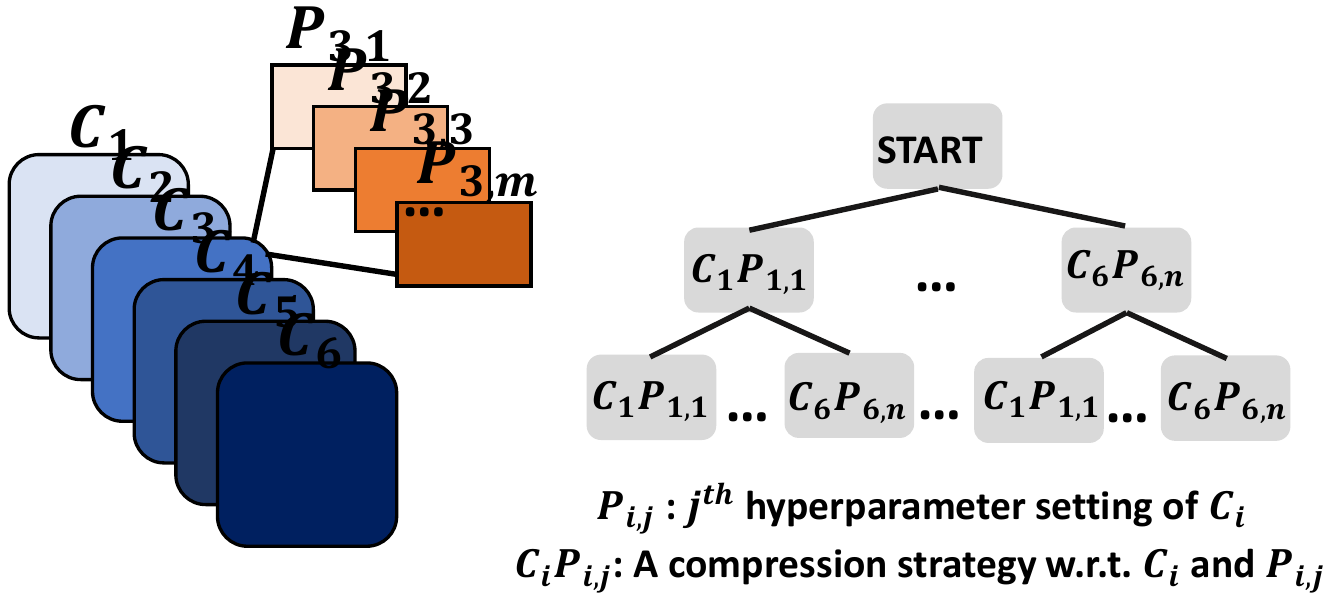}
\vspace{-0.3cm}
\caption{AutoMC's search space can be described in a tree structure. Each node has 4,525 children nodes, corresponding to the 4,525 compression strategies in Table~\ref{table1}.}
\label{fig1}
\vspace{-0.5cm}
\end{figure}

\subsection{Search Space on Compression Schemes}\label{section:3.2}
In AutoMC, we utilize some open source model compression methods to build a search space on model compression. Specifically, we collect 6 effective model compression methods, allowing them to be combined flexibly to obtain diverse model compression schemes to cope with different compression tasks. In addition, considering that hyperparameters have great impact on the performance of each method, we regard the compression method under different hyperparameter settings as different compression strategies, and intend to find the best compression strategy sequence, that is, the compression scheme, to effectively solve the actual compression problems.

\begin{figure*}[t]
\centering
\includegraphics[width=0.8\textwidth]{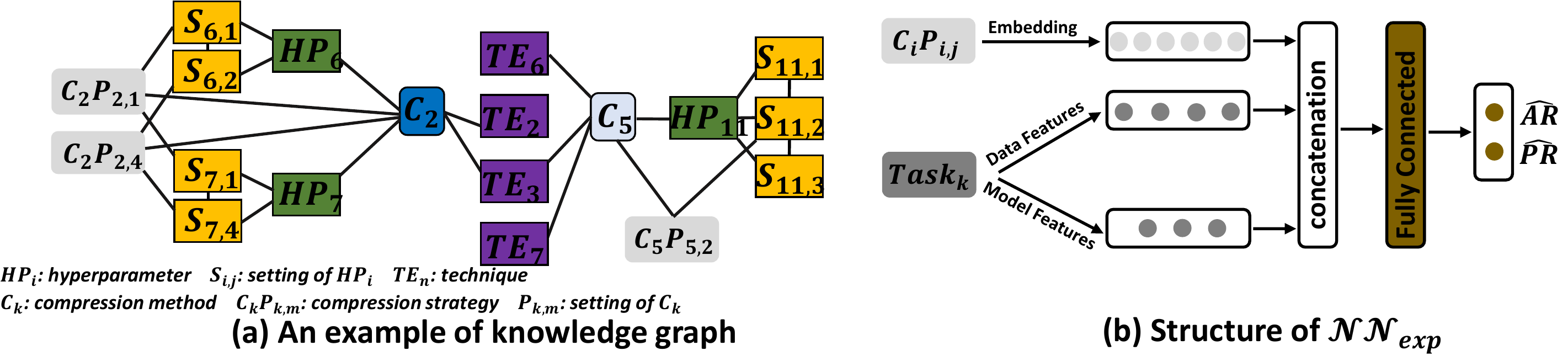}
\vspace{-0.3cm}
\caption{The structure of knowledge graph and $\mathcal{NN}_{exp}$ that are used for embedding learning. $S_{i,j}$ is the setting of hyperparameter $HP_{i}$.}
\label{fig2}
\vspace{-0.5cm}
\end{figure*}

Table~\ref{table1} gives these compression methods. These methods and their respective hyperparameters constitute a total of $4,525$ compression strategies. Utilizing these compression strategies to form compression strategy sequences of different lengths (length $< L$), then we get a search space $\mathbb{S}$ with $\sum_{l=0}^{L} (4525)^{l}$  different compression schemes.

Our search space $\mathbb{S}$ can be described as a tree structure  (as is shown in Figure~\ref{fig1}), where each node (layer $\leq L$) has $4,525$ child nodes corresponding to $4,525$ compression strategies and nodes at layer $L+1$ are leaf nodes. In this tree structure, each path from $START$ node to any node in the tree corresponds to a compression strategy sequence, namely a compression scheme in the search space. 

\subsection{Search Strategy of AutoMC Algorithm}\label{section:3.3}
The search space $\mathbb{S}$ is huge. In order to improve the search performance, we introduce domain knowledge to help AutoMC learn characteristics of components of $\mathbb{S}$ (Section~\ref{section:3.3.1}). In addition, we design a progressive search strategy to finely analyze the impact of subsequent operations on the compression scheme, and thus improve search efficiency (Section~\ref{section:3.3.2}). 

\subsubsection{Domain Knowledge based Embedding Learning}\label{section:3.3.1}
We build a knowledge graph on compression strategies, and extract experimental experience from the related research papers to learn potential advantages and effective representation of each compression strategy in the search space. Considering that two kinds of knowledge are of different types\footnote{knowledge graph is relational knowledge whereas experimental experience belongs to numerical knowledge} and are suitable for different analytical methods, we design different embedding learning methods for them, and combine two methods for better understanding of different compression strategies. 

\textbf{Knowledge Graph based Embedding Learning.} We build a knowledge graph $\mathbb{G}$ that exposes the technical and settings details of each compression strategy, to help AutoMC to learn relations and differences between different compression strategies. $\mathbb{G}$ contains five types of entity nodes: ($E_1$) compression strategy, ($E_2$) compression method, ($E_3$) hyperparameter, ($E_4$) hyperparameter’s setting and ($E_5$) compression technique. Also, it includes five types of entity relations:
\begin{itemize}
\small
\item[$R_1$:] corresponding relation between a compression strategy and its compression method ($E_1\to E_2$)
\vspace{-0.3cm}
\item[$R_2$:] corresponding relation between a compression strategy and its hyperparameter setting ($E_1\to E_4$)
\vspace{-0.3cm}
\item[$R_3$:] corresponding relation between a compression method and its hyperparameter ($E_2\to E_3$)
\vspace{-0.3cm}
\item[$R_4$:] corresponding relation between a compression method and its compression technique ($E_2\to E_5$ )
\vspace{-0.3cm}
\item[$R_5$:] corresponding relation between a hyperparameter and its setting ($E_3\to E_4$)
\end{itemize}
$R_1$ and $R_2$ describe the composition details of compression strategies, $R_3$ and $R_4$ provide a brief description of compression methods, $R_5$ illustrate the meaning of hyperparameter settings. Figure~\ref{fig2} (a) is an example of $\mathbb{G}$.

We use TransR~\cite{p1} to effectively parameterize entities and relations in $\mathbb{G}$ as vector representations, while preserving the graph structure of $\mathbb{G}$. Specifically, given a triplet $(h, r, t)$ in $\mathbb{G}$, we learn embedding of each entity and relation by optimizing the translation principle:
\begin{equation}
\small
W_{r} e_{h}+e_{r}\approx W_{r} e_{t}
\label{equ2}
\end{equation}
where $e_{h}, e_{t}\in R^{d}$ and $e_{r}\in R^{k}$ are the embedding for $h$, $t$, and $r$ respectively; $W_{r}\in R^{k\times d}$ is the transformation matrix of relation $r$. 

This embedding learning method can inject the knowledge  in $\mathbb{G}$ into representations of compression strategies, so as to learn effective representations of compression strategies. In AutoMC, we denote the embedding of compression strategy $C_{i} P_{i,j}$ learned from $\mathbb{G}$ by $e_{C_{i} P_{i,j}}$.

\begin{algorithm}[t]
	\small
	\caption{Compression Strategy Embedding Learning}
		\label{alg1}
		\centering
		\begin{algorithmic}[1]
		\STATE $\mathbb{C}\gets$ Compression strategies in Table~\ref{table1}
		\STATE $\mathbb{G}\gets$ Construct knowledge graph on $\mathbb{C}$
		\STATE $\mathbb{E}\gets$ Extract experiment experience w.r.t. $\mathbb{G}$ from papers involved in Table~\ref{table1}
		\WHILE {epoch $<$ $TrainEpoch$}
			\STATE Execute one epoch training of TransR using triplets in $\mathbb{G}$
			\STATE 	$e_{C_{i}P_{i,j}}\gets$ Extract knowledge embedding of compression strategy $C_{i}P_{i,j}$ $(\forall C_{i}P_{i,j}\in \mathbb{C})$
			\STATE 	Optimize the obtained knowledge embedding using $\mathbb{E}$ according to Equation~\ref{equ3}
			\STATE 	$\widetilde{e}_{C_{i}P_{i,j}}\gets$ Extract the enhanced embedding of $C_{i}P_{i,j}$ $(\forall C_{i}P_{i,j}\in \mathbb{C})$
			\STATE 	Replace $e_{C_{i}P_{i,j}}$ by $\widetilde{e}_{C_{i}P_{i,j}}$ $(\forall C_{i}P_{i,j}\in \mathbb{C})$
		\ENDWHILE
		\RETURN High-level embedding of compression strategies: $\widetilde{e}_{C_{i}P_{i,j}}$ $(\forall C_{i}P_{i,j}\in \mathbb{C})$
	    \end{algorithmic}
\end{algorithm}
\begin{figure}[t]
\centering
\vspace{-0.8cm}
\end{figure}

\textbf{Experimental Experience based Embedding Enhancement.} Research papers contain many valuable experimental experiences: the performance of compression strategies under a variety of compression tasks. These experiences are helpful for deeply understanding performance characteristics of each compression strategy. If we can integrate them into embeddings of compression strategies, then AutoMC can make more accurate decisions under the guidance of higher-quality embeddings.

Based on this idea, we design a neural network, which is denoted by $\mathcal{NN}_{exp}$ (as shown in Figure~\ref{fig2} (b)), to further optimize the embeddings of compression strategies learned from $\mathbb{G}$. $\mathcal{NN}_{exp}$ takes $e_{C_{i}P_{i,j}}$ and the feature vector of a compression task $Task_{k}$ (denoted by $e_{Task_{k}}$) as input, intending to output $C_{i}P_{i,j}$’s compression performance, including parameter's reduction rate $PR$, and accuracy’s increase rate $AR$, on $Task_{k}$. 

Here, $Task_{k}$ is composed of dataset attributes and model performance information. Taking the compression task on image classification model as an example, the feature vector can be composed of the following 7 parts: (1) Data Features: category number, image size, image channel number and data amount. (2) Model Features: original model’s parameter amount, FLOPs, accuracy score on the dataset.

In AutoMC, we extract experimental experience from relevant compression papers: $(C_{i} P_{i,j}, Task_{k},AR,PR)$, then input $e_{C_{i}P_{i,j}}$ and $e_{Task_{k}}$ to $\mathcal{NN}_{exp}$ to obtain the predicted performance scores, denoted by $(\hat{AR},\hat{PR})$. Finally, we optimize $e_{C_{i}P_{i,j}}$ and obtain a more effective embedding of $C_{i}P_{i,j}$, which is denoted by $\widetilde{e}_{C_{i}P_{i,j}}$, by minimizing the differences between $(AR,PR)$ and $(\hat{AR},\hat{PR})$:
\begin{equation}
\small
\begin{split}
\min\limits_{\theta,e_{C_{i}P_{i,j}} (C_{i}P_{i,j}\in \mathbb{C})} & \frac{1}{|\mathbb{E}|} \sum_{(C_{i}P_{i,j}, Task_{k},AR,PR)\in \mathbb{E}} \\ 
\lVert \mathcal{NN}_{exp}& \big(e_{C_{i} P_{i,j}},Task_{k};\theta\big)-(AR,PR)\rVert 
\end{split}
\label{equ3}
\end{equation}
where $\theta$ indicates the parameters of $\mathcal{NN}_{exp}$, $\mathbb{C}$ represents the set of compression strategies in Table~\ref{table1}, and $\mathbb{E}$ is the set of experimental experience extracted from papers.


\textbf{Pseudo code.} Combining the above two learning methods, then AutoMC can comprehensively consider knowledge graph and experimental experience and obtain a more effective embeddings. Algorithm~\ref{alg1} gives the complete pseudo code of the embedding learning part of AutoMC.

\subsubsection{Progressive Search Strategy}\label{section:3.3.2}

Taking the compression scheme as the unit to analyze and evaluate during the search phase can be very inefficient, since the compression scheme evaluation can be very expensive when its sequence is long. The search strategy may cost much time on evaluation while only obtain less performance information for optimization, which is ineffective. 

To improve search efficiency, we apply the idea of progressive search strategy instead in AutoMC. We try to gradually add the valuable compression strategy to the evaluated compression schemes by analyzing rich procedural information, i.e., the impact of each compression strategy on the original compression strategy sequence, so as to quickly find better schemes from the huge search space $\mathbb{S}$.

\begin{figure}[t]
\centering
\includegraphics[width=0.8\columnwidth]{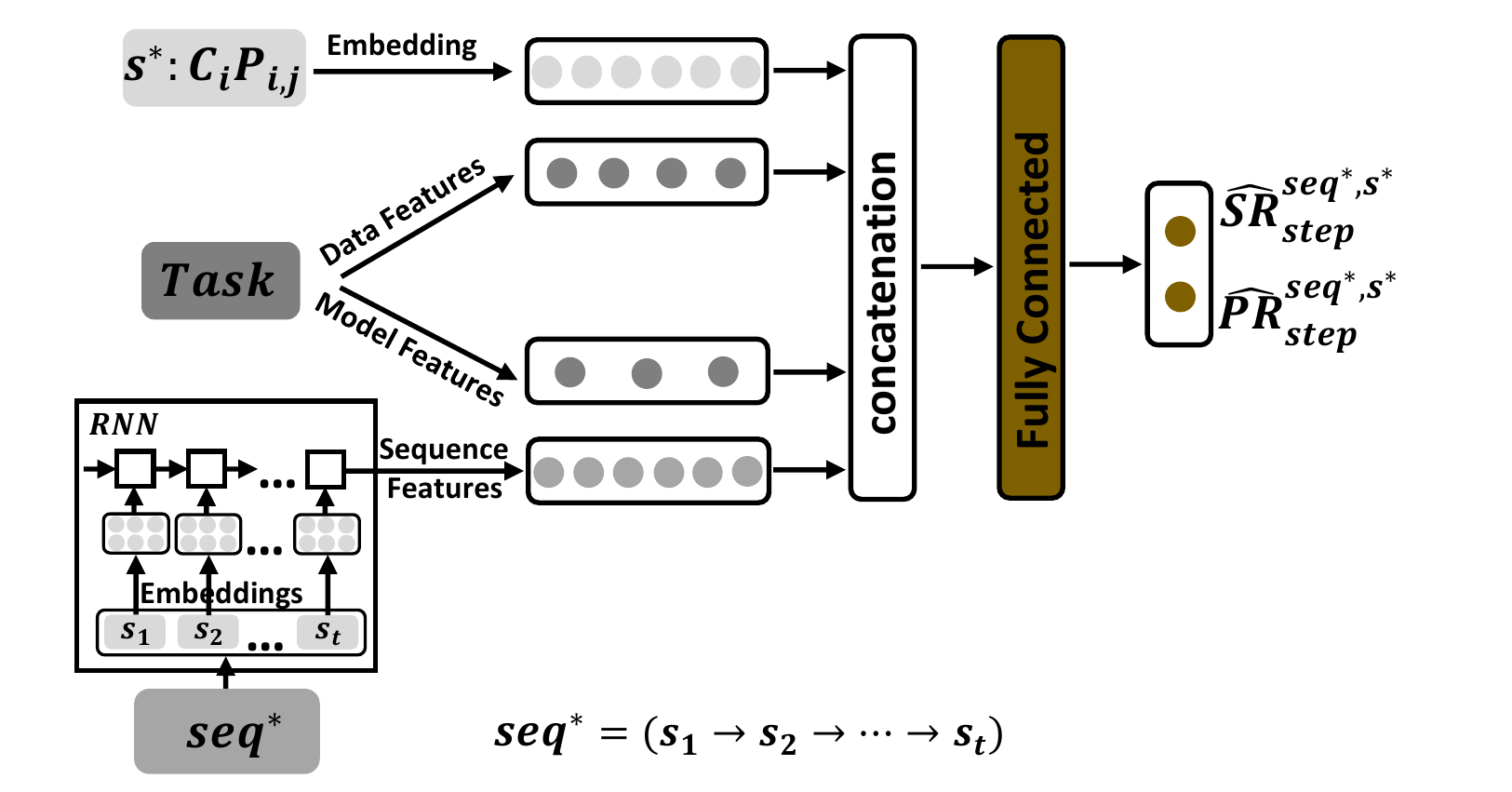}
\vspace{-0.2cm}
\caption{Structure of $\mathcal{F}_{mo}$. The embedding of $s_i$ and $s^{\ast}$ are provided by Algorithm~\ref{alg1}.}
\label{fig3}
\vspace{-0.5cm}
\end{figure}

Specifically, we propose to utilize historical procedural information to learn a multi-objective evaluator  $\mathcal{F}_{mo}$ (as shown in Figure~\ref{fig3}). We use $\mathcal{F}_{mo}$ to analyze the impact of a newly added compression strategy$ s_{t+1}=C_{i} P_{i,j}\in \mathbb{C}$ on the performance of compression scheme $seq=(s_1\to s_2\to \ldots \to s_t)$, including the accuracy improvement rate $AR_{step}$ and reduction rate of parameters $PR_{step}$.

\begin{algorithm}[t]
	\small
	\caption{Progressive Search Strategy}
		\label{alg2}
		\centering
		\begin{algorithmic}[1]
		\STATE $\mathcal{H}_{scheme}\gets \{START\}$, $OPT_{START}\gets \mathbb{C}$
		\WHILE {epoch $<$ $SearchEpoch$}
			\STATE $\mathcal{H}_{scheme}^{sub}\gets$ Sample some schemes from $\mathcal{H}_{scheme}$
			
			\STATE 	$\mathbb{S}_{step}\gets \{(seq,s)\ |\ \forall seq\in \mathcal{H}_{scheme}^{sub},\ s\in Next_{seq}\}$
			
			\STATE 	$ParetoO \gets \mathop{\argmax}_{(seq,s)\in \mathbb{S}_{step}} [ACC_{seq,s}, PAR_{seq,s}]$
			
			\STATE 	Evaluate schemes in $ParetoO$ and get $AR_{step}^{seq^{\ast},s^{\ast}}$, $PR_{step}^{seq^{\ast},s^{\ast}}$ $\big((seq^{\ast},s^{\ast})\in ParetoO\big)$
			\STATE 	Optimize the weights $\omega$ of multi-objective evaluator $\mathcal{F}_{mo}$ according to Equation~\ref{equ5}
			\STATE $\mathcal{H}_{scheme}\gets \mathcal{H}_{scheme}\cup \{seq^{\ast}, s^{\ast}\ |\ (seq^{\ast},s^{\ast})\in ParetoO)\}$
			\STATE $OPT_{seq^{\ast}}\gets OPT_{seq^{\ast}}-\{s^{\ast}\}$, $OPT_{seq^{\ast}\gets s^{\ast}}\gets \mathbb{C}$ for each $(seq^{\ast},s^{\ast})\in ParetoO$
			\STATE 	$ParetoSchemes\gets$ Pareto optimal compression schemes with parameter decline rate $\geq \gamma$ in $\mathcal{H}_{scheme}$
		\ENDWHILE
		\RETURN $ParetoSchemes$
	    \end{algorithmic}
\end{algorithm}

For each round of optimization, we firstly sample some Pareto-Optimal and evaluated schemes $seq\in \mathcal{H}_{scheme}$, take their next-step compression strategies $Next_{seq}\subseteq \mathbb{C}$ as the search space $\mathbb{S}_{step}$: $\mathbb{S}_{step}=\{(seq,s) | \forall seq\in \mathcal{H}_{scheme}^{sub}, s\in Next_{seq}\}$, where $\mathcal{H}_{scheme}^{sub}\subseteq \mathcal{H}_{scheme}$ are the sampled schemes. Secondly, use $\mathcal{F}_{mo}$ to select pareto optimal options $ParetoO$ from $\mathbb{S}_{step}$, thus obtain better compression schemes $seq^{\ast}\to s^{\ast}, \forall (seq^{\ast},s^{\ast})\in ParetoO$ for evaluation. 
\begin{equation}
\small
\begin{split}
ParetoO&=\mathop{\argmax}_{(seq,s)\in \mathbb{S}_{step}} [ACC_{seq,s}, PAR_{seq,s}] \\
ACC_{seq,s}&=A(seq[M])\times (1+\hat{AR}_{step}^{seq,s}) \\
PAR_{seq,s}&=P(seq[M])\times (1-\hat{PR}_{step}^{seq,s}) \\
\end{split}
\label{equ4}
\end{equation}
where $\hat{AR}_{step}^{seq,s}$ and $\hat{PR}_{step}^{seq,s}$ are performance changes that $s$ brings to scheme $seq$ predicted by $\mathcal{F}_{mo}$. $ACC_{seq,s}$  and $PAR_{seq,s}$ are accuracy and parameter amount obtained after executing scheme $seq\to s$ to the original model $M$. 

Finally, we evaluate compression schemes in $ParetoO$ and get their real performance changes, which are denoted by $AR_{step}^{seq^{\ast},s^{\ast}}$, $PR_{step}^{seq^{\ast},s^{\ast}}$, and use the following formula to further optimize the performance of $\mathcal{F}_{mo}$:
\begin{equation}
\small
\begin{split}
\min_\omega&  \frac{1}{|ParetoO|} \sum_{(seq^{\ast},s^{\ast})\in ParetoO} \\ 
&\Vert \mathcal{F}_{mo} (seq^{\ast},s^{\ast};\omega)-(AR_{step}^{seq^{\ast},s^{\ast}}, PR_{step}^{seq^{\ast},s^{\ast}})\Vert \\ 
\end{split}
\label{equ5}
\end{equation}

We add the new scheme $\{seq^{\ast}\to s^{\ast} | (seq^{\ast},s^{\ast})\in ParetoO)\}$ to $\mathcal{H}_{scheme}$ to participate in the next round of optimization steps.

\textbf{Advantages of Progressive Search and AutoMC.} In this way, AutoMC can obtain more training data for strategy optimization, and can selectively explore more valuable search space, thus improve the search efficiency. 

Applying embeddings learned by Algorithm~\ref{alg1} to Algorithm~\ref{alg2}, i.e., using the learned high-level embeddings to represent compression strategies and previous strategy sequences that need to input to $\mathcal{F}_{mo}$, then we get AutoMC.

\begin{table*}[t]
	\centering
	\caption{Compression results of ResNet-56 on CIFAR-10 and VGG-16 on CIFAR-100.}
	\vspace{-0.3cm}
	\resizebox{0.95\textwidth}{!}{
		\begin{tabular}{cc|ccc|ccc}
			\toprule[1pt]
			\multirow{2}{*}{\textbf{PR(\%)}} & \multirow{2}{*}{\textbf{Algorithm}} & \multicolumn{3}{c|}{\textbf{ResNet-56 on CIFAR-10}}                   & \multicolumn{3}{c}{\textbf{VGG-16 on CIFAR-100}}                      \\
			&                                     & Params(M) / PR(\%)    & FLOPs(G) / FR(\%)     & Acc. / Inc.(\%)       & Params(M) / PR(\%)    & FLOPs(G) / FR(\%)     & Acc. / Inc.(\%)       \\ \hline
			\multicolumn{1}{l}{}             & baseline                            & 0.90 / 0              & 0.27 / 0              & 91.04 / 0             & 14.77 / 0             & 0.63 / 0              & 70.03 / 0             \\ \hline
			\multirow{10}{*}{$\approx$ 40}   & LMA                                 & 0.53 / 41.74          & 0.15 / 42.93          & 79.61 / -12.56        & 8.85 / 40.11          & 0.38 / 40.26          & 42.11 / -39.87        \\
			& LeGR                                & 0.54 / 40.02          & 0.20 / 25.76          & 90.69 / -0.38         & 8.87 / 39.99          & 0.56 / 11.55          & 69.97 / -0.08         \\
			& NS                                  & 0.54 / 40.02          & 0.12 / 55.68          & 89.19 / -2.03         & 8.87 / 40.00          & 0.42 / 33.71          & 70.01 / -0.03         \\
			& SFP                                 & 0.55 / 38.52          & 0.17 / 36.54          & 88.24 / -3.07         & 8.90 / 39.73           & 0.38 / 39.31          & 69.62 / -0.58         \\
			& HOS                                 & 0.53 / 40.97          & 0.15 / 42.55          & 90.18 / -0.95         & 8.87 / 39.99          & 0.38 / 39.51          & 64.34 / -8.12         \\
			& LFB                                 & 0.54 / 40.19          & 0.14 / 46.12          & 89.99 / -1.15         & 9.40 / 36.21           & 0.04 / 93.00          & 60.94 / -13.04        \\
			\cline{2-8}
			& Evolution                           & 0.45 / 49.87          & 0.14 / 48.83          & 91.77 / 0.80          & 8.11 / 45.11          & 0.36 / 42.54          & 69.03 / -1.43         \\
			& \textbf{AutoMC}                        & \textbf{0.55 / 39.17} & \textbf{0.18 / 31.61} & \textbf{92.61 / 1.73} & \textbf{8.18 / 44.67} & \textbf{0.42 / 33.23} & \textbf{70.73 / 0.99} \\
			& RL                                  & 0.20 / 77.69          & 0.07 / 75.09          & 87.23 / -4.18         & 8.11 / 45.11          & 0.44 / 29.94          & 63.23 / -9.70         \\
			& Random                             & 0.22 / 75.95          & 0.06 / 77.18          & 79.50 / -12.43        & 8.10 / 45.15          & 0.33 / 47.80          & 68.45 / -2.25         \\ \hline
			\multirow{10}{*}{$\approx$ 70}   & LMA                                 & 0.27 / 70.40          & 0.08 / 72.09          & 75.25 / -17.35        & 4.44 / 69.98          & 0.19 / 69.90           & 41.51 / -40.73        \\
			& LeGR                                & 0.27 / 70.03          & 0.16 / 41.56          & 85.88 / -5.67         & 4.43 / 69.99          & 0.45 / 28.35          & 69.06 / -1.38         \\
			& NS                                  & 0.27 / 70.05          & 0.06 / 78.77          & 85.73 / -5.83         & 4.43 / 70.01          & 0.27 / 56.77          & 68.98 / -1.50         \\
			& SFP                                 & 0.29 / 68.07          & 0.09 / 67.24          & 86.94 / -4.51         & 4.47 / 69.72          & 0.19 / 69.22          & 68.15 / -2.68         \\
			& HOS                                 & 0.28 / 68.88          & 0.10 / 63.31          & 89.28 / -1.93         & 4.43 / 70.05          & 0.22 / 64.29          & 62.66 / -10.52        \\
			& LFB                                 & 0.27 / 70.03          & 0.08 / 71.96          & 90.35 / -0.76         & 6.27 / 57.44          & 0.03 / 95.2           & 57.88 / -17.35        \\
			\cline{2-8}
			& Evolution                           & 0.44 / 51.47          & 0.10 / 63.66          & 89.21 / -2.01         & 4.14 / 72.01          & 0.22 / 64.30          & 60.47 / -13.64        \\
			& \textbf{AutoMC}                        & \textbf{0.28 / 68.43} & \textbf{0.10 / 62.44} & \textbf{92.18 / 1.25} & \textbf{4.19 / 71.67} & \textbf{0.32 / 49.31} & \textbf{70.10 / 0.11} \\
			& RL                                  & 0.44 / 51.52          & 0.10 / 63.15          & 88.30 / -3.01         & 4.20 / 71.60          & 0.19 / 69.08          & 51.20 / -27.13        \\
			& Random                             & 0.43 / 51.98          & 0.13 / 52.53          & 88.36 / -2.94         & 5.03 / 65.94          & 0.28 / 55.37          & 51.76 / -25.87        \\ \bottomrule[1pt]
		\end{tabular}
	}
	\label{table-compressions}
	\vspace{-0.2cm}
\end{table*}

\begin{table*}[t]
	\centering
	\caption{Compression results of ResNets on CIFAR-10 and VGGs on CIFAR-100, setting target pruning rate as 40\%. Note that all data is formalized as PR(\%) / FR(\%) / Acc.(\%).}
	\vspace{-0.3cm}
	\resizebox{0.95\textwidth}{!}{
		\begin{tabular}{lllllll}
			\toprule[1pt]
			\textbf{Algorithm} & \textbf{ResNet-20 on CIFAR-10}         & \textbf{ResNet-56 on CIFAR-10}         & \textbf{ResNet-164 on CIFAR-10}        & \textbf{VGG-13 on CIFAR-100}           & \textbf{VGG-16 on CIFAR-100}           & \textbf{VGG-19 on CIFAR-100}           \\ \hline
			LMA                & 41.74 / 42.84 / 77.61          & 41.74 / 42.93 / 79.61          & 41.74 / 42.96 / 58.21          & 40.07 / 40.29 / 47.16          & 40.11 / 40.26 / 42.11          & 40.12 / 40.25 / 40.02          \\
			LeGR               & 39.86 / 21.20 / 89.20          & 40.02 / 25.76 / 90.69          & 39.99 / 33.11 / 83.93          & 40.00 / 12.15 / 70.80          & 39.99 / 11.55 / 69.97          & 39.99 / 11.66 / 69.64          \\
			NS                 & 40.05 / 44.12 / 88.78          & 40.02 / 55.68 / 89.19          & 39.98 / 51.13 / 83.84          & 40.01 / 31.19 / 70.48          & 40.00 / 33.71 / 70.01          & 40.00 / 41.34 / 69.34          \\
			SFP                & 38.30 / 35.49 / 87.81          & 38.52 / 36.54 / 88.24          & 38.58 / 36.88 / 82.06          & 39.68 / 39.16 / 70.69          & 39.73 / 39.31 / 69.62          & 39.76 / 39.40 / 69.42          \\
			HOS                & 40.12 / 39.66 / 88.81          & 40.97 / 42.55 / 90.18          & 41.16 / 43.50 / 84.12          & 40.06 / 39.36 / 64.13          & 39.99 / 39.51 / 64.34          & 40.01 / 39.13 / 63.37          \\
			LFB                & \textbf{40.38 / 45.80 / 91.57} & 40.19 / 46.12 / 89.99          & 40.09 / 76.76 / 24.17          & 37.82 / 92.92 / 63.04          & 36.21 / 93.00 / 60.94          & 35.46 / 93.05 / 56.27          \\
			\cline{1-7}
			Evolution          & 49.50 / 46.66 / 89.95          & 49.87 / 48.83 / 91.77          & 49.95 / 49.44 / 87.69          & 45.15 / 35.58 / 62.95          & 45.11 / 42.54 / 69.03          & 45.19 / 36.64 / 63.30          \\
			Random             & 75.94 / 74.44 / 78.38          & 75.95 / 77.18 / 79.50          & 75.91 / 78.08 / 59.37          & 45.18 / 24.04 / 62.02          & 45.15 / 47.80 / 68.45          & 45.11 / 33.06 / 68.81          \\
			RL                 & 77.87 / 69.05 / 84.28          & 77.69 / 75.09 / 87.23          & 77.23 / 83.27 / 74.21          & 45.20 / 26.00 / 62.36          & 45.11 / 29.94 / 63.23          & 45.14 / 38.78 / 68.31          \\
			\textbf{AutoMC}    & 38.73 / 30.00 / 91.42          & \textbf{39.17 / 31.61 / 92.61} & \textbf{39.30 / 40.76 / 88.50} & \textbf{44.60 / 34.43 / 71.77} & \textbf{44.67 / 33.23 / 70.73} & \textbf{44.68 / 35.09 / 70.56} \\ \bottomrule[1pt]
		\end{tabular}
	}
	\label{table-tansfer}
	\vspace{-0.2cm}
\end{table*}

\section{Experiments}\label{section:4}
In this part, we examine the performance of AutoMC. We firstly compare AutoMC with human designed compression methods to analyze AutoMC’s application value and the rationality of its search space design (Section~\ref{section:4.3}). Secondly, we compare AutoMC with classical AutoML algorithms to test the effectiveness of its search strategy (Section~\ref{section:4.4}). Then, we transfer the compression scheme searched by AutoMC to other neural models to examine its transferability (Section~\ref{section:4.5}). Finally, we conduct ablation studies to analyze the impact of embedded learning method based on domain knowledge and progressive search strategy on the overall performance of AutoMC (Section~\ref{section:4.6}).

We implemented all algorithms using Pytorch and performed all experiments using RTX 3090 GPUs.

\subsection{Experimental Setup }\label{section:4.1}
\textbf{Compared Algorithms.} We compare AutoMC with two popular search strategies for AutoML: a RL search strategy that combines recurrent neural network controller~\cite{p2} and EA-based search strategy for multi-objective optimization~\cite{p2}, and a commonly used baseline in AutoML, Random Search. To enable these AutoML algorithms to cope with our automatic model compression problem, we set their search space to $\mathbb{S}$ ($L=5$). In addition, we take 6 state-of-the-art human-invented compression methods: LMA~\cite{C1}, LeGR~\cite{C2}, NS~\cite{C3}, SFP~\cite{C4}, HOS~\cite{C5} and LFB~\cite{C7}, as baselines, to show the importance of automatic model compression. 

\begin{figure*}[t] 
	\centering
	\subfloat[Achieved highest accuracy score (Exp1)]{
		\begin{minipage}[t]{0.24\linewidth}
			\centering 
			\includegraphics[width=1\columnwidth]{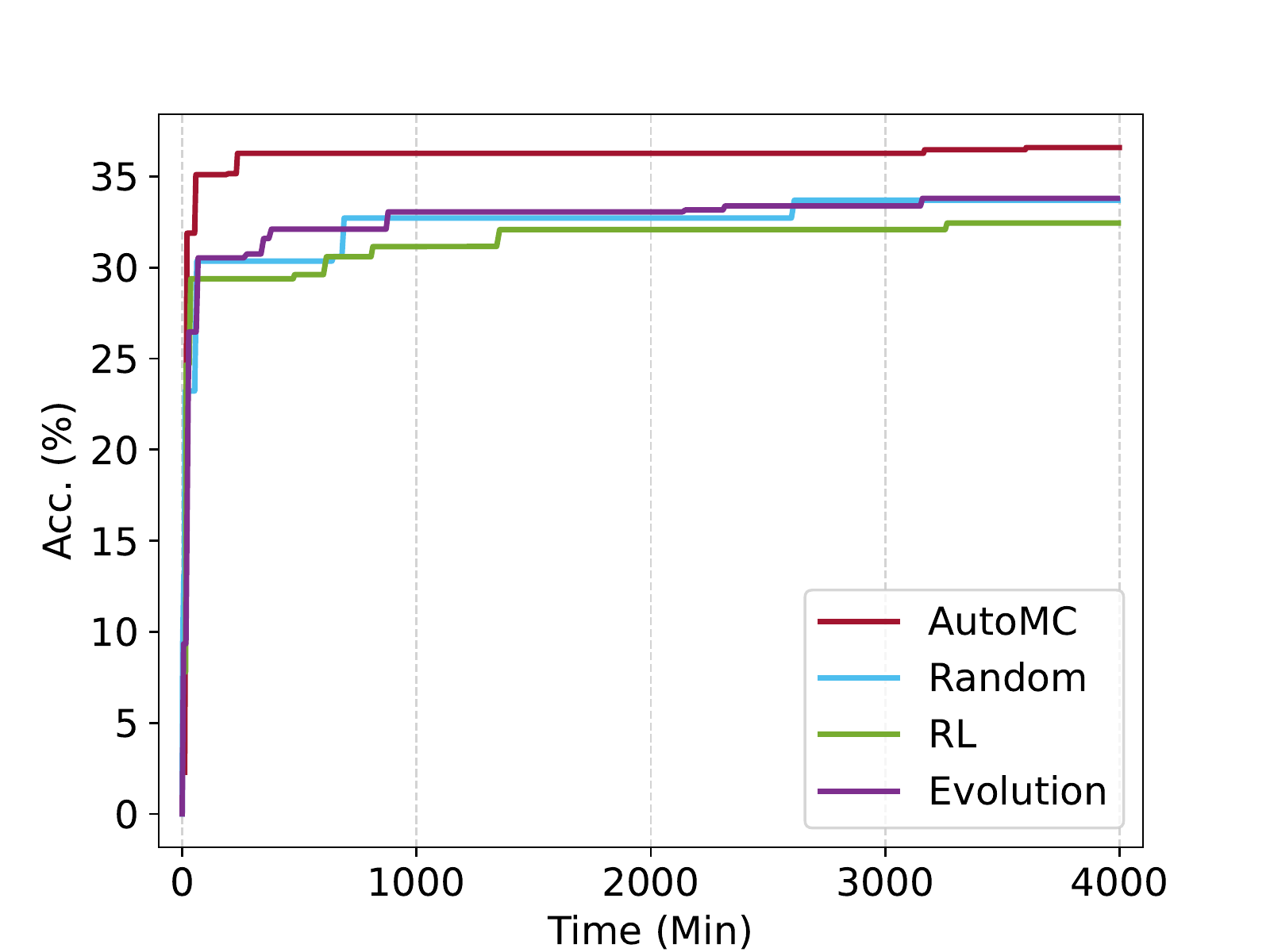}
		\end{minipage}
	}
	\subfloat[Final Pareto front (Exp1)]{
		\begin{minipage}[t]{0.24\linewidth}
			\centering   
			\includegraphics[width=1\columnwidth]{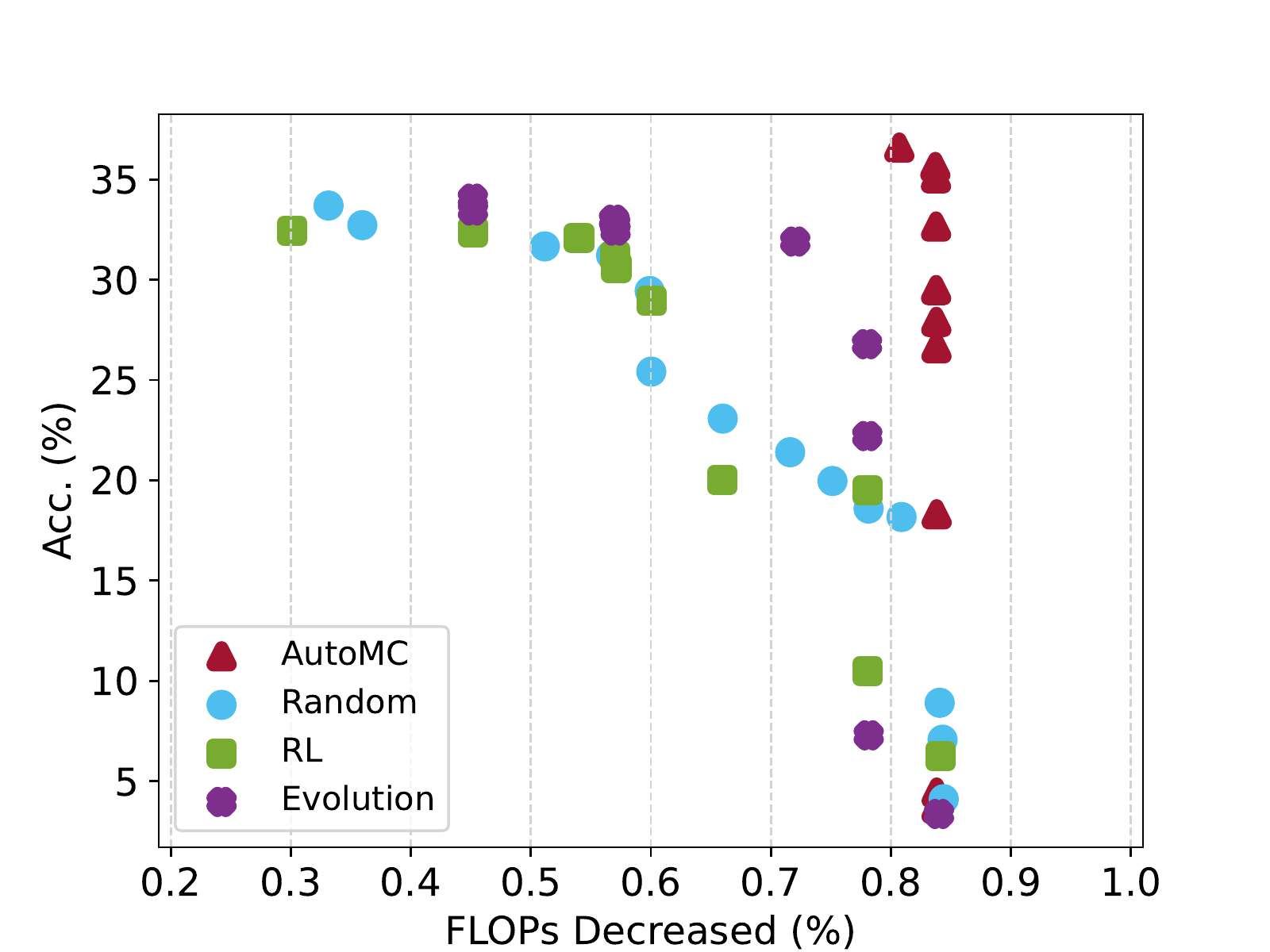} 
		\end{minipage}
	}
	\subfloat[Achieved highest accuracy score (Exp2)]{
		\begin{minipage}[t]{0.24\linewidth}
			\centering 
			\includegraphics[width=1\columnwidth]{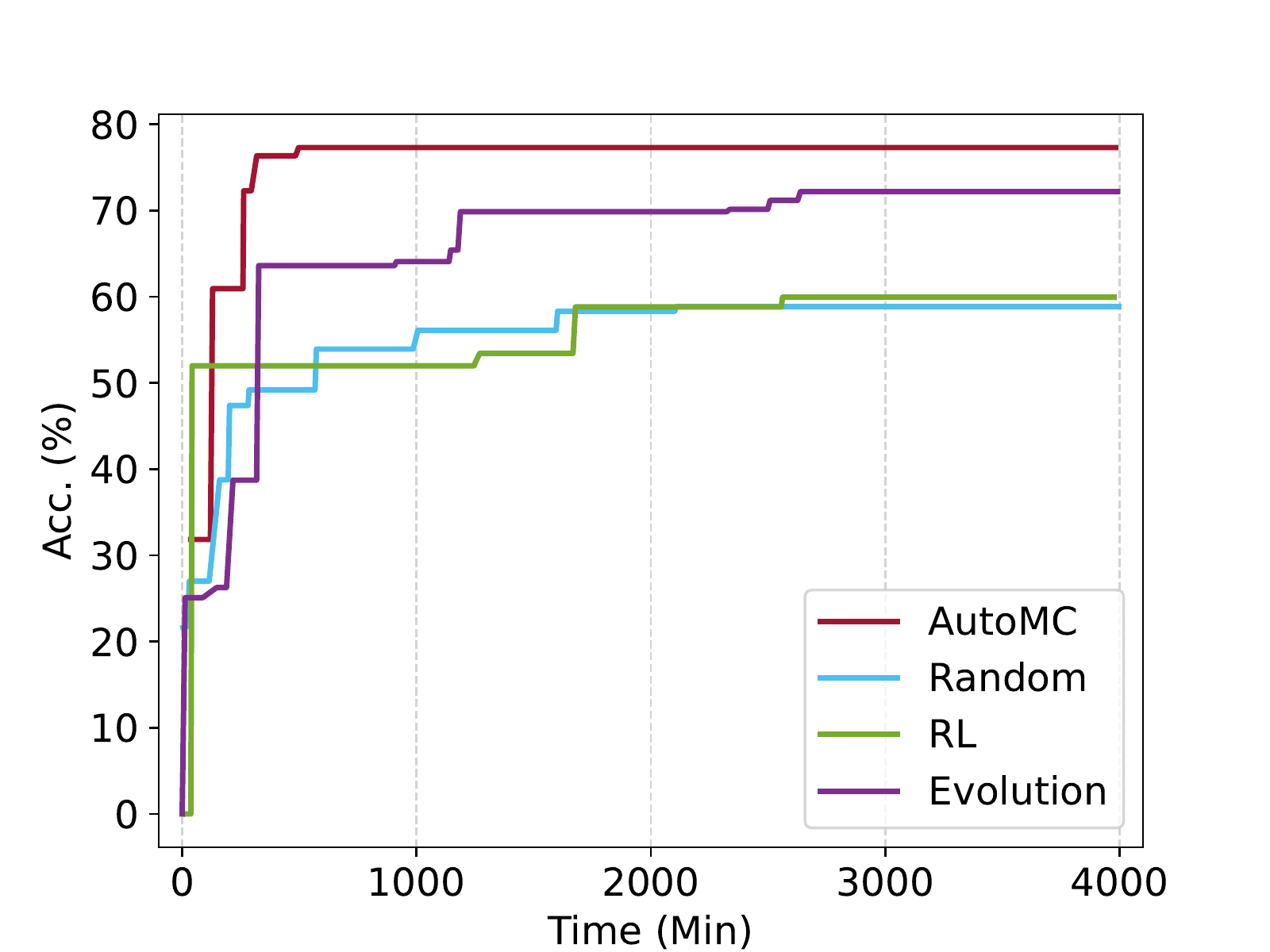}
		\end{minipage}
	}
	\subfloat[Final Pareto front (Exp2)]{
		\begin{minipage}[t]{0.24\linewidth}
			\centering   
			\includegraphics[width=1\columnwidth]{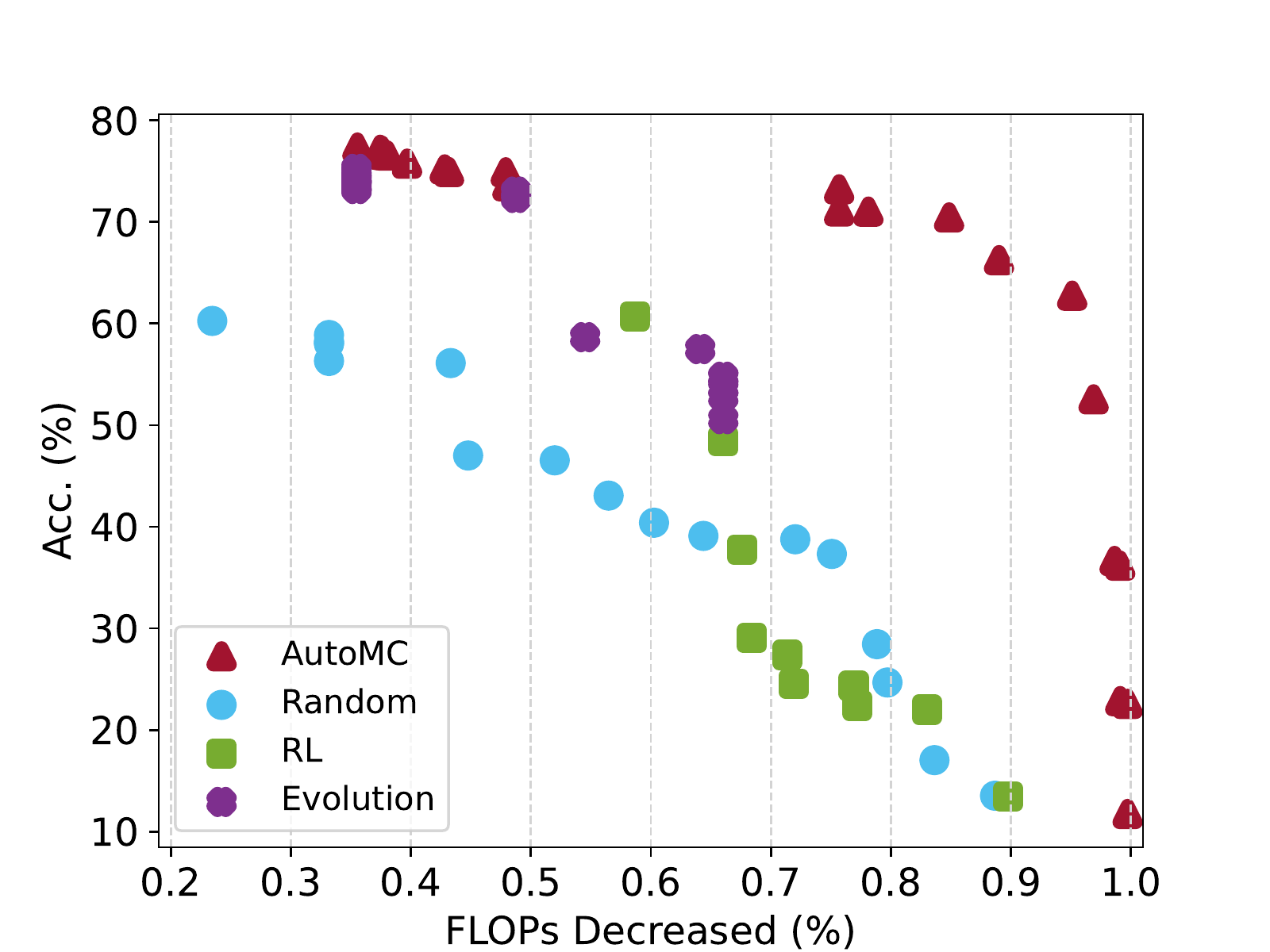} 
		\end{minipage}
	}
	\vspace{-0.3cm}
	\caption{Pareto optimal results searched by different AutoML algorithms on Exp1 and Exp2. } 
	\label{fig4}
	\vspace{-0.3cm}
\end{figure*}

\textbf{Compression Tasks.} We construct two experiments to examine the performance of AutoML algorithms. \textbf{Exp1:} $D$=CIAFR-10, $M$= ResNet-56, $\gamma$=0.3;  \textbf{Exp2:} $D$= CIAFR-100, $M$=VGG-16, $\gamma$=0.3, where CIAFR-10 and CIAFR-100~\cite{cifar} are two commonly used image classification datasets, and ResNet-56 and VGG-16 are two popular CNN network architecture. 

To improve the execution speed, we sample 10\% data from $D$ to execute AutoML algorithms in the experiments. After executing AutoML algorithms, we select the Pareto optimal compression scheme with $PR\geq \gamma$ for evaluation. As for the existing compression methods, we apply grid search to get their optimal hyperparameter settings and set their parameter reduction rate to 0.4 and 0.7 to analyze their compression performance.

Furthermore, to evaluate the transferability of compression schemes searched by AutoML algorithms, we design two transfer experiments. We transfer compression schemes searched on ResNet-56 to ResNet-20 and ResNet-164, and transfer schemes from VGG-16 to VGG-13 and VGG-19.

\textbf{Implementation Details.}  In AutoMC, the embedding size is set to $32$. $\mathcal{NN}_{exp}$ and $\mathcal{F}_{mo}$ are trained with the Adam with a learning rate of 0.001. After AutoMC searches for 3 GPU days, we choose the Pareto optimal compression schemes as the final output. As for the compared AutoML algorithms, we follow implementation details reported in their papers, and control the running time of each AutoML algorithm to be the same. Figure~\ref{fig6} gives the best compression schemes searched by AutoMC.

\subsection{Comparison with the Compression Methods}\label{section:4.3} 
Table \ref{table-compressions} gives the performance of AutoMC and the existing compression methods on different tasks. We can observe that compression schemes designed by AutoMC surpass the manually designed schemes in all tasks. These results prove that AutoMC has great application value. It has the ability to help users search for better compression schemes automatically to solve specific compression tasks.

In addition, the experimental results show us: (1) A compression strategy may performs better with smaller parameter reduction rate ($PR$). Taking result of ResNet-56 on CIFAR-10 using LeGR as an example, when the $PR$ is 0.4, on average, the model performance falls by 0.0088\% for every 1\% fall in parameter amount; however, when $PR$ becomes larger, the model performance falls by 0.0737\% for every 1\% fall in parameter amount. (2) Different compression strategies may be appropriate for different compression tasks. For example, LeGR performs better than HOS when the $PR=0.4$ whereas HOS outperforms LeGR when $PR=0.7$. Based on the above two points, combination of multiple compression strategies and fine-grained compression for a given compression task may achieve better results. This is consistent with our idea of designing the AutoMC search space, and it further proves the rationality of the AutoMC search space design.

\subsection{Comparison with the NAS algorithms}\label{section:4.4} 
Table \ref{table-compressions} gives the performance of different AutoML algorithms on different compression tasks. Figure~\ref{fig4} provides the performance of the best compression scheme (Pareto optimal scheme with highest accuracy score) and all Pareto optimal schemes searched by AutoML algorithms. We can observe that RL algorithm performs well in the very early stage, but its performance improvement is far behind other AutoML algorithms in the later stage. Evolution algorithm outperforms the other algorithms except AutoMC in both experiments. As for the Random algorithm, its performance have been rising throughout the entire process, but still worse than most algorithms. Compared with the existing AutoML algorithms, AutoMC can search for better model compression schemes more quickly, and is more suitable for the search space which contains a huge number of candidates. These results demonstrate the effectiveness of AutoMC and the rationality of its search strategy design.


\begin{figure*}[t] 
	\centering
	\subfloat[Achieved highest accuracy score (Exp1)]{
		\begin{minipage}[t]{0.24\linewidth}
			\centering 
			\includegraphics[width=\columnwidth]{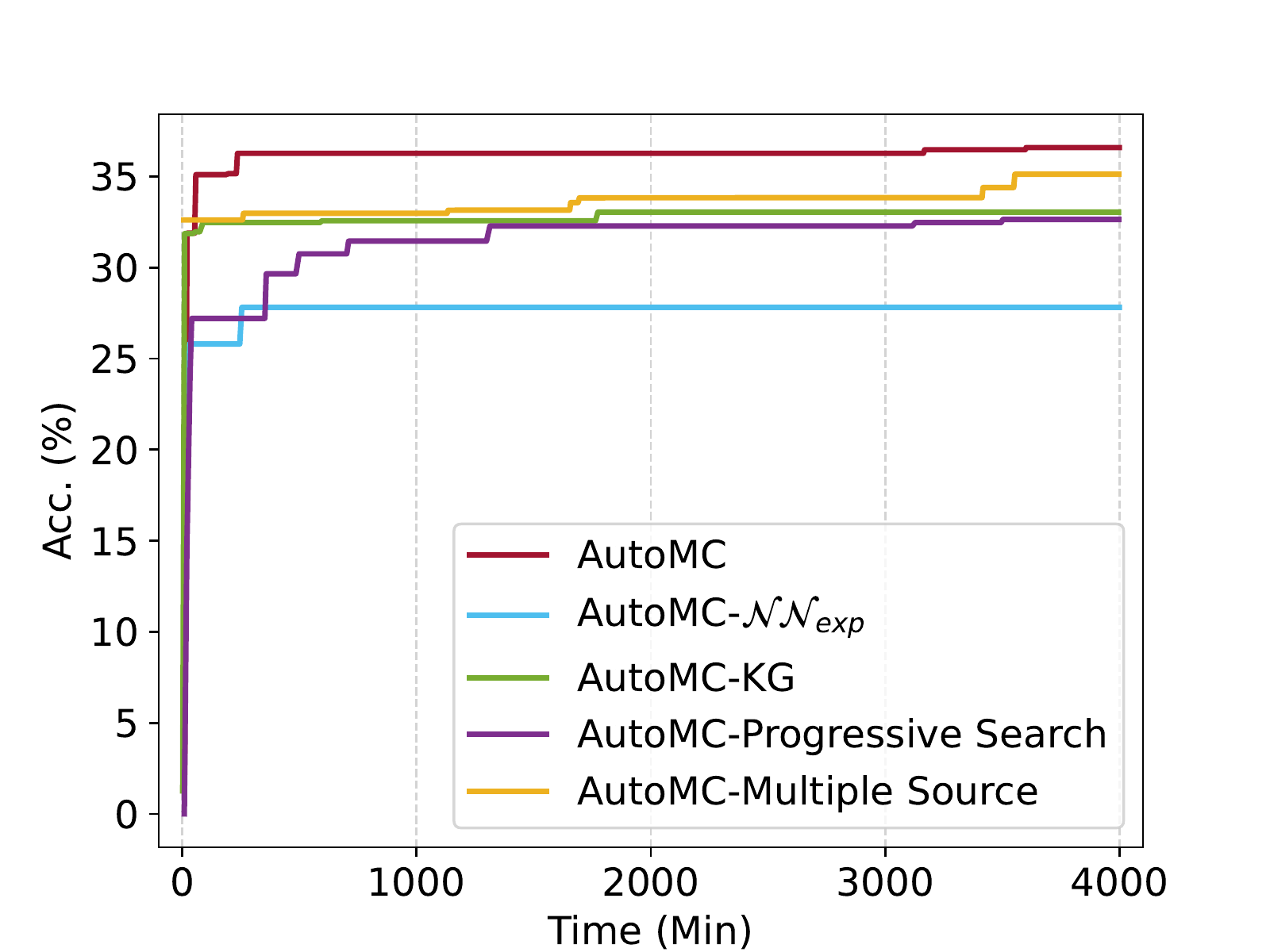}
		\end{minipage}
	}
	\subfloat[Final Pareto front (Exp1)]{
		\begin{minipage}[t]{0.24\linewidth}
			\centering   
			\includegraphics[width=\columnwidth]{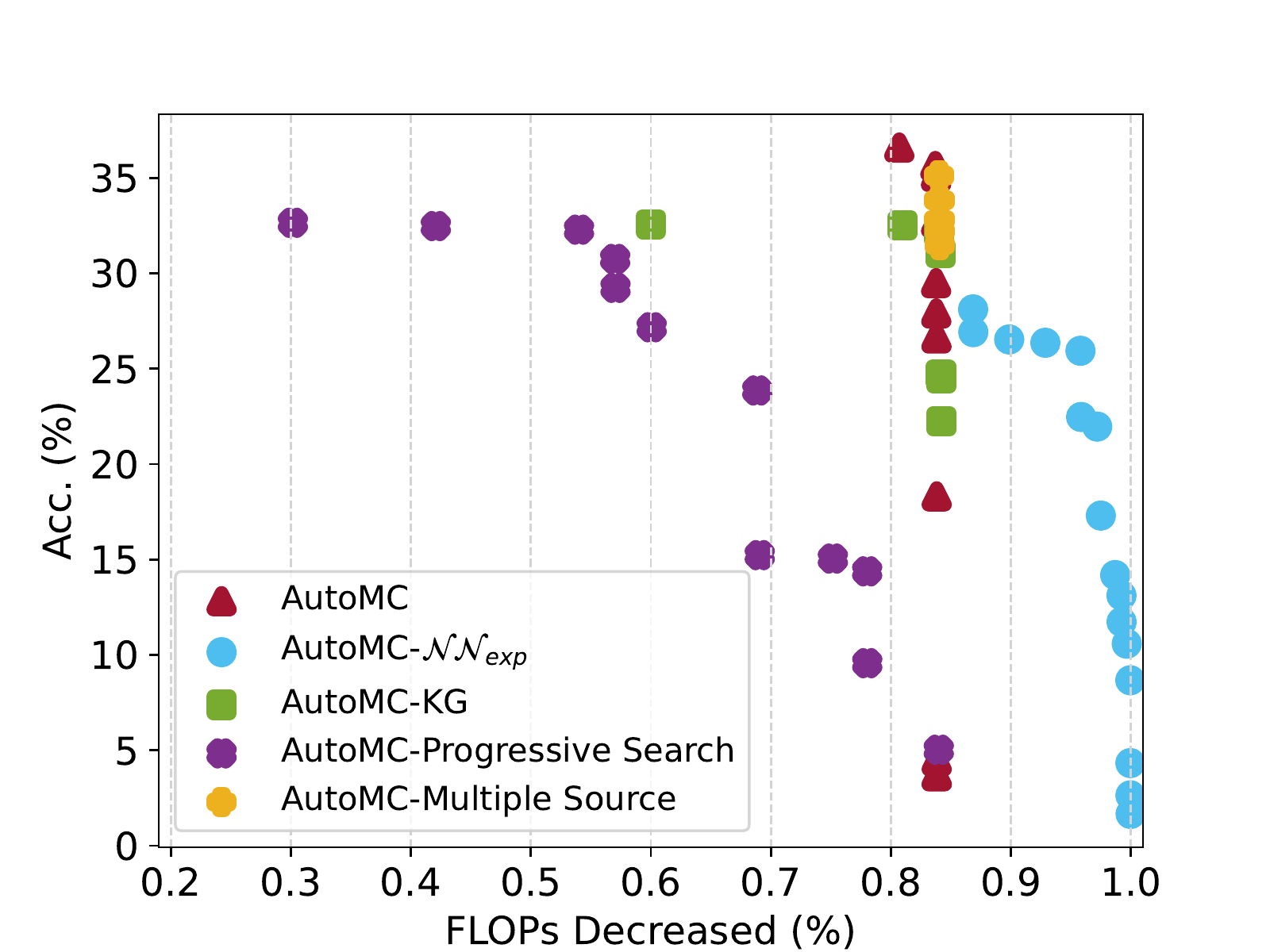} 
		\end{minipage}
	}
	\subfloat[Achieved highest accuracy score (Exp2)]{
		\begin{minipage}[t]{0.24\linewidth}
			\centering 
			\includegraphics[width=\columnwidth]{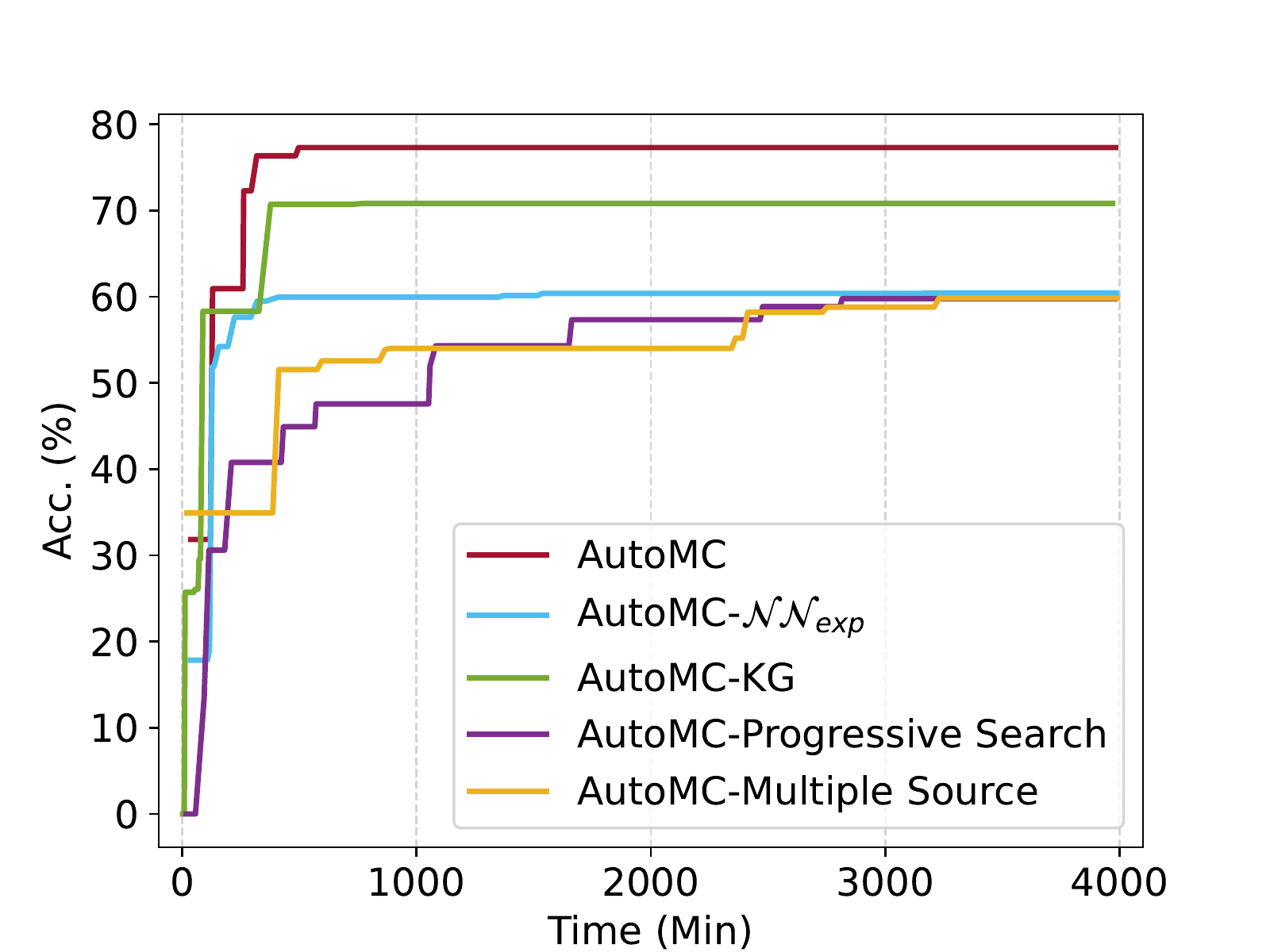}
		\end{minipage}
	}
	\subfloat[Final Pareto front (Exp2)]{
		\begin{minipage}[t]{0.24\linewidth}
			\centering   
			\includegraphics[width=\columnwidth]{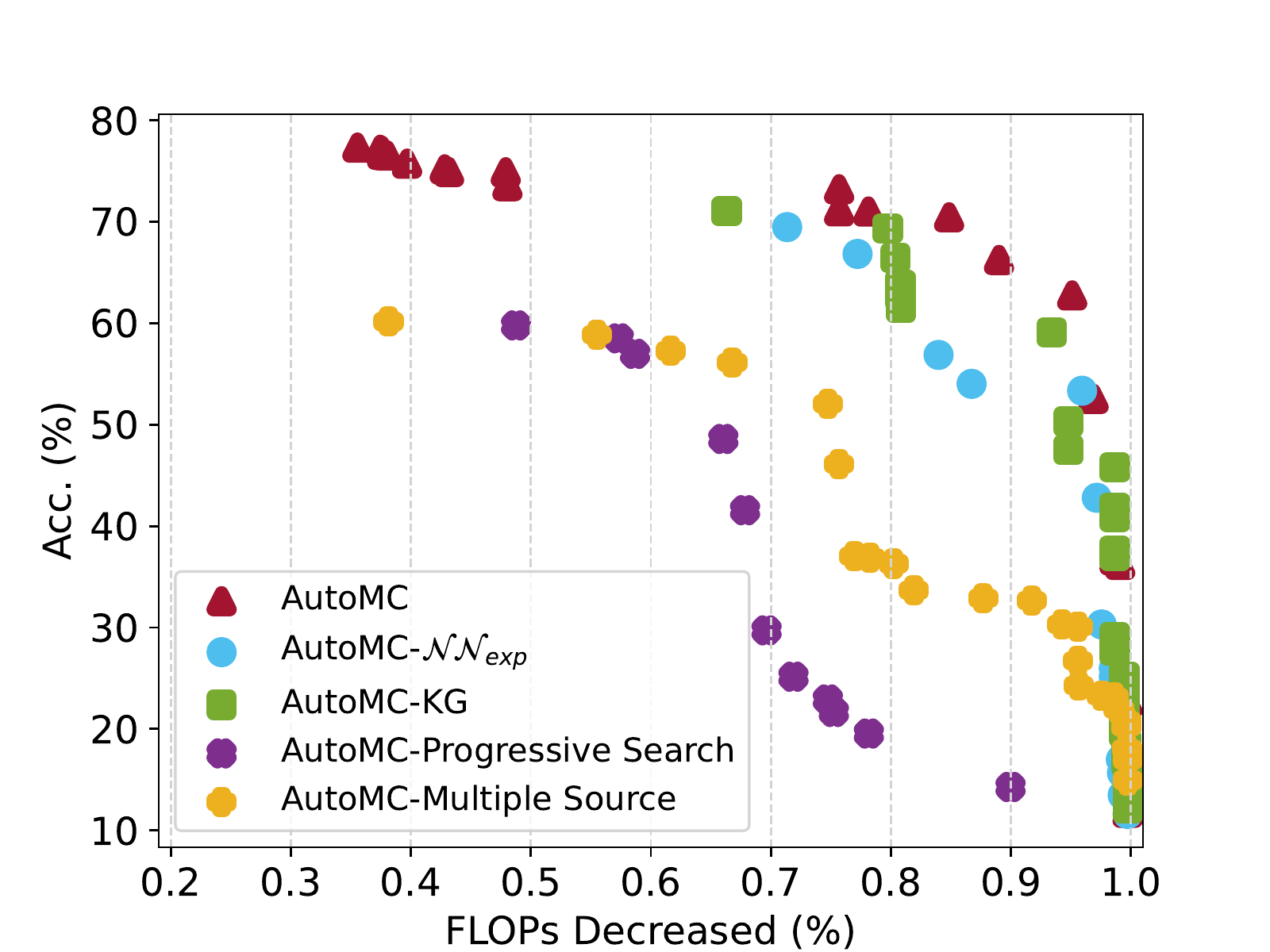} 
		\end{minipage}
	}
	\vspace{-0.3cm}
	\caption{Pareto optimal results serach by different versions of AutoMC on Exp1 and Exp2.} 
	\label{fig_alg}
	\vspace{-0.4cm}
\end{figure*}

\begin{figure}[t]
	\centering
	\includegraphics[width=0.9\linewidth]{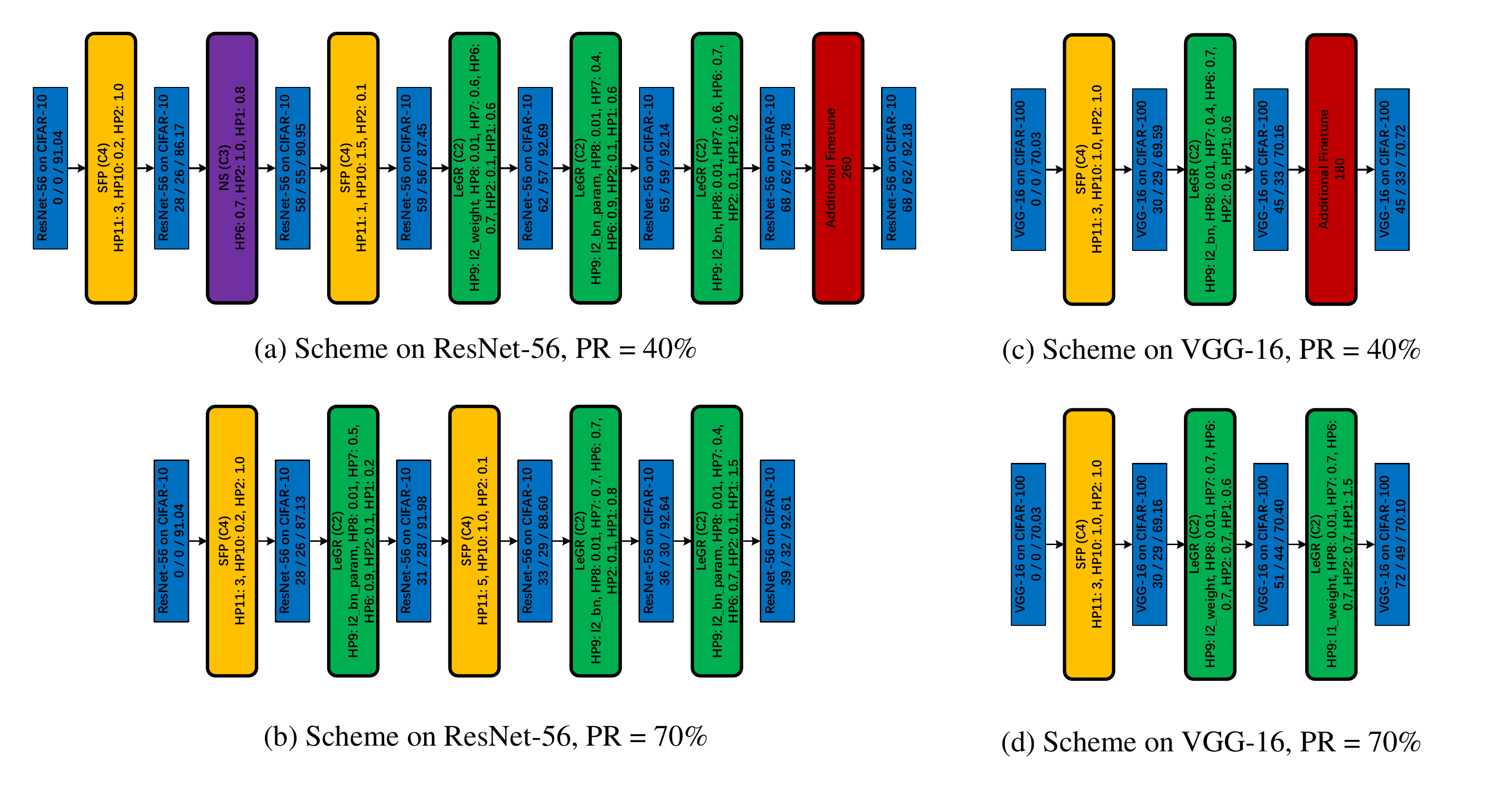}
	\vspace{-0.55cm}
	\caption{The compression schemes searched by AutoMC. Addtional fine-tuning will be added to the end of sequence to make up fine-tuning epoch for comparison.}
	\label{fig6}
	\vspace{-0.5cm}
\end{figure}

\subsection{Tansfer Study}\label{section:4.5} 
Table \ref{table-tansfer} shows the performance of different models transfered from ResNet-56 and VGG-16. We can observe that LFB outperforms AutoMC with ResNet-20 on CIFAR-10. We think the reason is that LFB has a talent for dealing with small models. It's obvious that the performance of LFB gradually decreases as the scale of the model increases. For example, LFB achieves an accuracy of 91.57\% with ResNet-20 on CIFAR-10, but only achieves 24.17\% with ResNet-164 on CIFAR-10. Except that, compression schemes designed by AutoMC surpass the manually designed schemes in all tasks. These results prove that AutoMC has great transferability. It is able to help users search for better compression schemes automatically with models of different scales.

Besides, the experimental results show that the same compression strategies may achieve diferent performance on models of different scales. In addition to the example of LFB and AutoMC above, LeGR performs better than HOS when using ResNet-20 whereas HOS outperforms LeGR when using ResNet-164. Based the above, combination of multiple compression strategies and fine-grained compression for models of different scales may achieve more stable and competitive performance.

\subsection{Ablation Study}\label{section:4.6} 
We further investigate the effect of the knowledge based embedding learning method, experience based embedding learning method and the progressive search strategy, three core components of our algorithm, on the performance of AutoMC using the following four variants of AutoMC, thus verify innovations presented in this paper. 
\begin{itemize}
\small
\item[1] \textit{AutoMC-KG}. This version of AutoMC removes knowledge graph embedding method.
\vspace{-0.3cm}
\item[2] \textit{AutoMC-$\mathcal{NN}_{exp}$}. This version of AutoMC removes experimental experience based embedding method.
\vspace{-0.3cm}
\item[3] \textit{AutoMC-Multiple Source}. This version of AutoMC only uses strategies w.r.t. LeGR to construct search space. 
\vspace{-0.3cm}
\item[4] \textit{AutoMC-Pregressive Search}. This version of AutoMC replaces the progressive search strategy with the RL based search strategy that combines recurrent neural network. 
\end{itemize}
Corresponding results are shown in Figure~\ref{fig_alg}, we can see that AutoMC has much better performance than \textit{AutoMC-KG} and \textit{AutoMC-$\mathcal{NN}_{exp}$}, which ignore the knowledge graph or experimental experience on compression strategies while learning their embedding. This result shows us the significance and necessity of fully considering two kinds of knowledge on compression strategies in the AutoMC, for effective embedding learning. Our proposed knowledge graph embedding method can explore the differences and linkages between compression strategies in the search space, and the experimental experience based embedding method can reveal the performance characteristics of compression strategies. Two embedding learning methods can complement each other and help AutoMC have a better and more comprehensive understanding of search space components. 

Also, We notice that \textit{AutoMC-Multiple Source} achieve worse performance than AutoMC. \textit{AutoMC-Multiple} use only one compression method to complete compression tasks. The result indicates the importance of using multi-source compression strategies to build the search space.

Besides, we observe that \textit{AutoMC-Progressive Search} performs much worse than AutoMC. RL’s unprogressive search process, i.e., only search for, evaluate, and analyze complete compression schemes, performs worse in the automatic compression scheme design problem task. It fails to effectively use historical evaluation details to improve the search effect and thus be less effective than AutoMC.

\section{Conclusion}\label{section:5}
In this paper, we propose the AutoMC to automatically design optimal compression schemes according to the requirements of users. AutoMC innovatively introduces domain knowledge to assist search strategy to deeply understand the potential characteristics and advantages of each compression strategy, so as to design compression scheme more reasonably and easily. In addition, AutoMC presents the idea of progressive search space expansion, which can selectively explore valuable search regions and gradually improve the quality of the searched scheme through finer-grained analysis. This strategy can reduce the useless evaluations and improve the search efficiency. Extensive experimental results show that the combination of  existing compression methods can create more powerful compression schemes, and the above two innovations make AutoMC more efficient than existing AutoML methods. In future works, we will try to enrich our search space, and design a more efficient search strategy to tackle this search space for further improving the performance of AutoMC.

{\small
\bibliographystyle{ieee_fullname}
\bibliography{egbib}
}
\end{document}